\documentclass{article} % For LaTeX2e
\usepackage[ruled,vlined]{algorithm2e}
\usepackage{iclr2027_conference,times}

\usepackage{amsmath,amsfonts,bm}

\def\eqref#1{equation~\ref{#1}}
\def\1{\bm{1}}

\DeclareMathAlphabet{\mathsfit}{\encodingdefault}{\sfdefault}{m}{sl}
\SetMathAlphabet{\mathsfit}{bold}{\encodingdefault}{\sfdefault}{bx}{n}

\newcommand{\softmax}{\mathrm{softmax}}

\usepackage{booktabs}
\usepackage{graphicx}

\title{HERO-MoE: Historical Expert Routing \\ with Scale-Preserving Fusion}
\author{
\textbf{Junxiang Qiu}\textsuperscript{1},
\textbf{Zhengsu Chen}\textsuperscript{2},
\textbf{Xinting Hu}\textsuperscript{1,*},
\textbf{Shuo Wang}\textsuperscript{1},
\textbf{Hengheng Zhang}\textsuperscript{2}\\[2pt]
\textbf{Shaofeng Zhang}\textsuperscript{1},
\textbf{Changcheng Li}\textsuperscript{1},
\textbf{Boyu Shi}\textsuperscript{3},
\textbf{Qi Tian}\textsuperscript{2,*}\\[4pt]
\normalfont
\textsuperscript{1}University of Science and Technology of China
\quad
\textsuperscript{2}Huawei Inc.
\quad
\textsuperscript{3}Southeast University\\[3pt]
\texttt{qiujx@mail.ustc.edu.cn},
\quad
\texttt{xinting@ustc.edu.cn}
\quad
\texttt{tian.qi1@huawei.com}\\[3pt]
\textsuperscript{*}Corresponding authors
}

\iclrfinalcopy % Uncomment for camera-ready version, but NOT for submission.
\begin{document}

\maketitle

\begin{abstract}
Mixture-of-Experts (MoE) architectures have become a standard way to scale model capacity while keeping computation sparse, yet routing remains a key determinant of MoE quality and training behavior. Prior empirical studies suggest that MoE routing reflects input semantics and upstream computation across depth, but standard routers do not explicitly use the routing distributions produced by preceding layers. We propose \textbf{HERO-MoE}, \textbf{H}istorical \textbf{E}xpert \textbf{RO}uting with Scale-Preserving Fusion, a routing framework that injects historical routing priors into MoE routers by reusing detached, dense routing distributions collected from preceding MoE layers. The key idea is simple: HERO-MoE preserves the original token-conditioned routing branch and adds a residual historical routing contribution before the standard softmax and top-$k$ dispatch. To stabilize this historical signal, HERO-MoE introduces a scale-preserving fusion mechanism that matches the magnitude of historical routing memory to the current hidden representation and accounts for the number of visible historical layers, without introducing an auxiliary routing loss or a fusion-specific tuning parameter. By reusing routing distributions already computed by preceding MoE layers, HERO-MoE improves training-loss reduction with modest end-to-end overhead. The resulting router remains compatible with standard sparse dispatch, including top-$k$ and group-limited routing, and can be inserted into existing MoE backbones with minimal architectural changes. Experiments on an approximately 8B-parameter MoE model with 0.5B active parameters, trained from scratch on 100B tokens, show that HERO-MoE reduces the final loss from 1.6393 to 1.6184, while peak memory and FLOPs increase by only 0.44\% and 0.64\%, respectively.

% Mixture-of-Experts（MoE）架构已经成为在保持计算稀疏的同时扩展模型容量的标准方法，但路由机制仍然是影响 MoE 模型质量和训练行为的关键因素。已有实证研究表明，MoE 路由能够反映输入语义以及随深度传播的上游计算信息，但标准 router 并不会显式利用前序层产生的路由分布。我们提出 \textbf{HERO-MoE}，即带有尺度保持融合的历史专家路由方法（\textbf{H}istorical \textbf{E}xpert \textbf{RO}uting with Scale-Preserving Fusion）。HERO-MoE 通过复用前序 MoE 层产生的、与计算图分离的稠密路由分布，将历史路由先验注入当前 MoE router。其核心思想很简单：HERO-MoE 保留原始的基于当前 token 的路由分支，并在标准 softmax 和 top-$k$ 调度之前加入一个历史路由产生的残差贡献。为了稳定这一历史信号，HERO-MoE 引入尺度保持融合机制，使历史路由记忆的幅值与当前 hidden representation 保持匹配，并考虑可见历史层数量的影响，同时不引入辅助 routing loss 或 fusion-specific 调参参数。通过复用前序 MoE 层已经计算得到的路由分布，HERO-MoE 能够以较小的端到端开销改善训练 loss 的下降。所得 router 仍然兼容标准稀疏调度流程，包括 top-$k$ 和 group-limited routing，并且可以通过较小的结构改动插入现有 MoE 主干。基于一个总参数量约 8B、激活参数量约 0.5B 的 MoE 模型进行实验，该模型从随机初始化开始在 100B tokens 上预训练。结果表明，HERO-MoE 将最终 loss 从 1.6393 降低到 1.6184，同时 peak memory 仅比 baseline 增加 0.44\%。

\end{abstract}

\section{Introduction}
Mixture-of-Experts (MoE) \citep{liu2024deepseekv3,zhou2022mixture} has become a widely adopted paradigm for scaling large language models, because it increases model capacity without increasing the computation of every token proportionally. By routing each token to only a small subset of experts, MoE architectures achieve a favorable trade-off between parameter scale and computational efficiency, and have demonstrated strong performance in language modeling, translation, and long-context reasoning. From early sparsely gated models to modern large-scale sparse Transformers, the central mechanism has remained the same: a router selects a small number of experts for each token, thereby determining how expert specialization emerges and how model capacity is utilized \citep{jacobs1991adaptive,shazeer2017outrageously,lepikhin2020gshard,liu2024deepseek,fedus2022switch,du2022glam,zoph2022st}. Consequently, the routing mechanism has become one of the most important components affecting the quality and optimization behavior of sparse MoE models.
% Mixture-of-Experts（MoE）已经成为扩展大语言模型的一种重要范式，因为它能够在不按比例增加每个 token 计算量的情况下提升模型容量。通过将每个 token 只路由到少量专家，MoE 架构能够在参数规模和计算效率之间取得较好的平衡，并已在语言建模、机器翻译和长上下文推理等任务中展现出较强性能。从早期的稀疏门控模型到现代大规模稀疏 Transformer，其核心机制始终一致：router 为每个 token 选择少量专家，从而决定专家专门化如何形成以及模型容量如何被利用。因此，routing 机制已经成为影响稀疏 MoE 模型质量和优化行为的关键组件之一。

Prior empirical studies consistently show that MoE routing reflects input semantics and upstream computation, with structured routing patterns and cross-layer effects extending across depth~\citep{olson2025probing,muennighoff2025olmoe,li2026understanding}. These findings indicate that preceding routing distributions contain token-specific and computation-aware signals that can complement the current hidden representation. Although a standard router computes logits from the current hidden representation, which carries information from preceding computation, it does not explicitly take the routing distributions produced by preceding layers as additional inputs. Treating routing history as an explicit prior can therefore provide the current router with a direct signal about the same token's previous expert assignments, enabling later expert selection to use both current token features and the token's preceding routing trajectory.
% 以往实证研究一致表明，MoE 路由反映了输入语义和上游计算过程，其结构化的路由模式以及跨层影响能够延伸到不同深度~\citep{olson2025probing,muennighoff2025olmoe,li2026understanding}。这些研究说明，前序层产生的路由分布包含依赖于 token 且与计算过程相关的信息，可以对当前 hidden representation 形成补充。尽管标准 router 根据当前 hidden representation 计算路由 logits，而该表示本身携带了前序计算的信息，但它并不会显式接收前序层产生的路由分布作为额外输入。因此，将 routing history 作为显式先验，可以为当前 router 提供关于同一 token 在前序层中专家选择的直接信息，使后续专家选择能够同时利用当前 token 特征和该 token 已经形成的历史路由轨迹.
Existing work has begun to exploit related routing structure through indirect regularization or by modifying the routing pathway. Regularization-based methods encourage consistent, non-overlapping, or complementary expert assignments~\citep{omi2025load,guo2026advancing,hu2026synergistic}, whereas context-aware routing methods share, propagate, or recurrently update routing-related states across layers~\citep{qiu2025layerwise,gu2025omni,huang2026multi}. However, these approaches either require additional regularization terms, loss weights, and method-specific hyperparameter tuning, or modify the routing pathway itself. Directly injecting preceding routing distributions as historical priors while preserving the original current-token routing branch and standard sparse dispatch therefore remains underexplored. This leaves an important design gap in how to exploit routing history without sacrificing the simplicity and compatibility of the standard MoE routing pipeline.
% 现有工作已经开始通过间接正则化或修改路由路径来利用相关的路由结构。基于正则化的方法鼓励专家分配保持一致、减少重叠或在不同深度上形成互补\citep{omi2025load,guo2026advancing,hu2026synergistic}；上下文感知路由方法则通过在层间共享、传播或循环更新与路由相关的状态来利用跨层信息\citep{qiu2025layerwise,gu2025omni,huang2026multi}。然而，这些方法要么需要额外的正则化项、损失权重以及方法相关的超参数调节，要么会修改路由路径本身。因此，在保留原始当前 token 路由分支和标准稀疏调度流程的同时，将前序路由分布作为历史先验直接注入当前 router，仍然缺乏充分研究。这形成了一个重要的设计空白：如何利用 routing history，同时不牺牲标准 MoE 路由流程的简洁性与兼容性。

To fill this gap, we propose HERO-MoE, Historical Expert Routing with Scale-Preserving Fusion. The key idea is simple: HERO-MoE reuses the detached, dense routing distributions produced by preceding MoE layers within the same pipeline-parallel stage~\citep{huang2019gpipe} as historical priors for the current router. It preserves the original token-conditioned routing branch and adds a residual logit contribution derived from the routing history before the standard softmax and top-$k$ dispatch. Thus, the router can exploit how the same token was routed in preceding layers without changing the standard sparse dispatch procedure or introducing an additional training objective. The main challenge lies in the scale of this seemingly simple fusion. As the number of visible historical layers grows, directly combining unnormalized routing history can make its contribution either too weak or too dominant, while also causing layer-dependent variations in the router input. HERO-MoE addresses this issue with scale-preserving normalization and depth-aware initialization, which align the historical and current representations and maintain a comparable initial logit scale across layers. Since the historical distributions are detached from the computation graph, the method introduces modest additional computation and does not backpropagate through previous routing decisions.
% 为填补这一空白，我们提出了 HERO-MoE，即带有尺度保持融合的历史专家路由方法。其核心思想很简单：HERO-MoE 复用同一流水线并行阶段内前序 MoE 层产生的、与计算图分离的稠密路由分布~\citep{huang2019gpipe}，并将其作为当前 router 的历史先验。该方法保留原有的基于 token 的路由分支，并在标准 softmax 和 top-$k$ 调度之前，加入由历史路由信息产生的残差 logits。因此，当前 router 可以利用同一 token 在前序层中的路由行为，同时无需改变标准稀疏调度流程或引入额外的训练目标。
% 然而，这一看似简单的融合方式对尺度控制提出了关键要求。随着可见历史层数量增加，直接融合未经归一化的路由历史可能导致其贡献过弱或过强，并造成不同层之间 router 输入尺度的变化。为此，HERO-MoE 引入尺度保持的归一化机制和深度感知初始化，使历史信息与当前表示的尺度保持匹配，并维持不同层之间可比的初始 logit 尺度。由于历史路由分布与计算图分离，该方法不会对前序路由决策进行反向传播，并且只引入可以忽略的额外计算开销。

We evaluate HERO-MoE on an approximately 8B-parameter MoE model with 0.5B active parameters trained from scratch on 100B tokens. HERO-MoE reduces the final training loss from 1.6393 to 1.6184 and improves the average score from 36.57 to 37.42 across 13 general benchmarks, with improvements on 9 tasks. The additional end-to-end cost remains modest: compared with the baseline, HERO-MoE increases total step FLOPs by only 0.64\%, peak memory by 0.44\%, and total training time by 3.85\%. These results show that historical routing can improve both MoE optimization and downstream generalization without requiring an auxiliary routing loss or additional method-specific tuning. Furthermore, the learned historical router provides an interpretable view of layer-wise routing dependencies and expert-level cooperative or competitive relations.
% 我们在一个约 8B 总参数、0.5B 激活参数的 MoE 模型上评估 HERO-MoE，该模型从零开始在 100B tokens 上训练。HERO-MoE 将最终训练损失从 1.6393 降低到 1.6184，并在 13 个通用 benchmark 上将平均分从 36.57 提升到 37.42，其中 9 个任务取得提升。额外的端到端开销较小：相比 baseline，HERO-MoE 的 total step FLOPs 仅增加 0.64%，peak memory 增加 0.44%，总训练时间增加 3.85%。这些结果表明，历史 routing 能够在不引入辅助 routing loss 或方法相关调参的情况下，同时改善 MoE 的优化过程和下游泛化性能。此外，训练得到的历史 router 还提供了一个可解释视角，用于观察层间 routing 依赖以及专家级协同或竞争关系。
In summary, our contributions are threefold:
\begin{itemize}
    \item We identify and exploit non-trivial cross-layer dependencies in MoE routing by proposing HERO-MoE, a historical expert routing framework that injects stage-local routing priors into the current router.

    \item We introduce a scale-preserving historical fusion mechanism with depth-aware initialization, which stabilizes the magnitude of historical routing signals without requiring an additional fusion-specific hyperparameter or auxiliary routing loss.
    
    \item We demonstrate that HERO-MoE improves training loss and downstream benchmark performance with modest end-to-end overhead, while also providing an interpretable visualization of cross-layer and inter-expert routing relationships.
\end{itemize}

% 总而言之，我们的贡献包括以下三点：

% 我们发现并利用了 MoE routing 中非平凡的跨层依赖关系，提出了 HERO-MoE，一种将 stage-local routing priors 注入当前 router 的历史专家路由框架。

% 我们引入了带有 depth-aware initialization 的 scale-preserving historical fusion 机制，该机制能够稳定历史 routing 信号的幅值，同时不需要额外的 fusion-specific 超参数或辅助 routing loss。

% 我们证明 HERO-MoE 能够在较小端到端开销下改善训练损失和下游 benchmark 表现，同时还能够提供对跨层和专家间 routing 关系的可解释可视化。

\section{Related Work}
\label{sec:related_work}
In this section, we review studies on MoE routing along two related threads: sparse routing with load balancing and specialization regularization, and context-aware system-level routing designs.
\subsection{Sparse MoE Routing and Load Balancing}
Mixture-of-Experts research dates back to early conditional computation and hierarchical expert models \citep{jacobs1991adaptive,jordan1994hierarchical}, and was later scaled to modern neural networks with sparsely gated routing \citep{shazeer2017outrageously}. Subsequent work such as GShard, Switch Transformers, GLaM, and ST-MoE demonstrated that sparse top-$k$ routing can substantially improve parameter efficiency and training scalability in large language models \citep{lepikhin2020gshard,fedus2022switch,du2022glam,zoph2022st}. However, these routers still rely primarily on the current token representation when making dispatch decisions, which makes them vulnerable to expert collapse, uneven utilization, and unstable training dynamics. To mitigate these issues, a number of balancing strategies were proposed, including expert-choice routing and auxiliary-loss-free balancing \citep{zhou2022mixture,wang2024auxiliary}. These methods improve load balance, but they still treat routing as a mostly local decision problem and do not explicitly model how routing behavior evolves across depth.
% 中文：MoE 研究可以追溯到早期的条件计算和层次化专家模型，随后通过稀疏门控路由扩展到现代神经网络。后续工作如 GShard、Switch Transformers、GLaM 和 ST-MoE 表明，稀疏 top-$k$ 路由能够显著提升大语言模型中的参数效率和训练可扩展性。然而，这些路由器在进行派发决策时仍然主要依赖当前 token 表示，因此容易出现专家塌缩、利用不均和训练动态不稳定等问题。为了缓解这些问题，已有研究提出了多种平衡策略，包括 expert-choice routing 和 auxiliary-loss-free balancing。这些方法改善了负载均衡，但它们仍然把路由主要视为一个局部决策问题，并没有显式建模路由行为如何随深度演化。

Recent work also shows that auxiliary regularization losses can improve specialization and routing consistency without changing the router architecture. Load Balancing Mixture of Experts with Similarity Preserving Routers introduces a similarity-preserving load-balancing loss that encourages consistent expert choices for similar inputs \citep{omi2025load}. Advancing Expert Specialization for Better MoE adds orthogonality and variance losses to reduce expert overlap and sharpen routing decisions \citep{guo2026advancing}. Synergistic Intra- and Cross-Layer Regularization Losses for MoE Expert Specialization combines intra-layer specialization and cross-layer coupling losses to promote complementary expert usage across depth \citep{hu2026synergistic}. These methods are effective, but they incur additional loss weights and hyperparameter tuning overhead; furthermore, the optimal parameters of each individual method require extra adjustment when different methods operate in conjunction, whereas HERO-MOE improves routing mainly by injecting historical routing priors through a scale-preserving router parameterization.
% 中文：近期工作还表明，在不改变路由器架构的情况下，辅助正则化损失也可以提升专家专门化和路由一致性。Load Balancing Mixture of Experts with Similarity Preserving Routers 引入了一种相似性保持的负载均衡损失，鼓励相似输入选择一致的专家。Advancing Expert Specialization for Better MoE 通过加入正交损失和方差损失来减少专家重叠并增强路由决策。Synergistic Intra- and Cross-Layer Regularization Losses for MoE Expert Specialization 则结合层内专门化和跨层耦合损失，以促进不同深度上的互补专家使用。这些方法是有效的，但也会增加额外的损失权重和调参开销；相比之下，HERO-MOE 主要通过保尺度的路由器参数化注入历史路由先验来改进路由。

\subsection{Context-Aware and System-Aware Routing}
A growing line of work argues that MoE routing should exploit more than the current hidden state alone. RMoE introduces layerwise recurrent routing and shows that cross-layer dependencies can improve expert selection and parameter efficiency \citep{qiu2025layerwise}. Omni-Router shares routing decisions across MoE layers to encourage structured expert usage in speech models \citep{gu2025omni}, while MCF-MOE incorporates multi-level context to obtain more consistent expert selection across layers \citep{huang2026multi}. Complementary analyses on shared routing geometry and similarity-preserving routers further suggest that routers learn reusable relational structure, and that preserving such structure can reduce redundancy and improve balancing \citep{labzin2026evidence,omi2025load,ahrac2026routers}. In parallel, system-oriented MoE frameworks such as DeepSpeed-MoE, FastMoE, Tutel, and MegaBlocks improve training and inference efficiency through better parallelism, dispatch, and kernel design \citep{rajbhandari2022deepspeed,he2021fastmoe,hwang2023tutel,gale2023megablocks}. Unlike these methods, HERO-MoE integrates historical routing context directly into the router while preserving the standard sparse dispatch procedure, making the routing policy more aware of cross-layer routing structure.
% 中文：越来越多的研究认为，MoE 路由不应只利用当前隐状态。RMoE 引入了逐层递归路由，并表明跨层依赖可以提升专家选择和参数效率。Omni-Router 在 MoE 层之间共享路由决策，以鼓励语音模型中的结构化专家使用；MCF-MOE 则引入多层级上下文，以获得更一致的跨层专家选择。关于共享路由几何和相似性保持路由器的补充分析进一步表明，路由器会学习可复用的关系结构，而保持这种结构能够减少冗余并改善均衡性。与此同时，DeepSpeed-MoE、FastMoE、Tutel 和 MegaBlocks 等系统导向的 MoE 框架通过更好的并行、派发和 kernel 设计提升训练与推理效率。不同于这些方法，HERO-MOE 将历史上下文直接融入路由器，同时平衡专家、设备和通信负载，使路由策略本身更加 stage 感知和系统感知。

\section{Method}
\label{sec:method}

In this section, we formalize the complete HERO-MoE framework. We first provide an overview of the routing pipeline, and then describe the router parameterization, scale-preserving initialization strategy, and stage-local historical routing mechanism in detail.

\begin{figure*}[t]
  \centering
  \includegraphics[width=\linewidth]{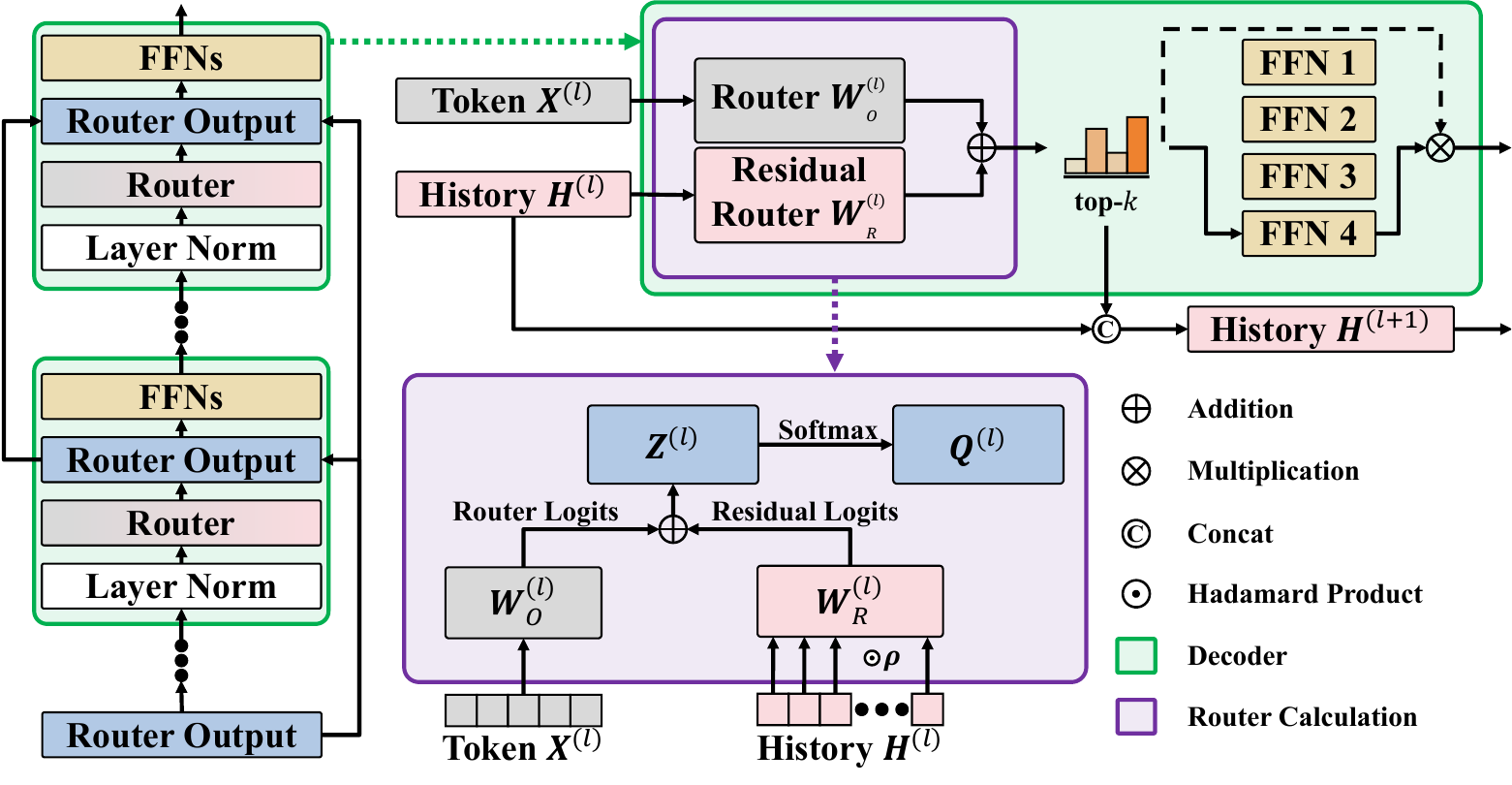}
  \caption{
Pipeline of HERO-MoE, including the original router branch, historical residual branch, sparse dispatch, and history update.
}
  \label{fig:pipeline}
\end{figure*}

Figure~\ref{fig:pipeline} illustrates the overall routing flow of HERO-MoE. For each MoE layer, the current token representation is processed by the original router branch, while the historical routing memory collected from preceding layers is fed into the residual router branch. The two branches produce original router logits and residual historical logits, which are added to obtain the final router logits. The resulting logits are converted into routing probabilities for top-$k$ expert selection and weighted expert aggregation. Meanwhile, the dense routing output of the current layer is concatenated with the previous history to form the updated historical memory for subsequent MoE layers.

\subsection{Parameterization}

Consider a Transformer with $L$ MoE layers. Let $\bm{X}^{(l)} \in \mathbb{R}^{N \times d}$ denote the token hidden representations entering the $l$-th MoE layer, where $N$ is the sequence length or batch token count, $d$ is the hidden dimension, $l \in \{1,\dots,L\}$ is the layer index, $L$ is the total number of MoE layers, and $E$ is the total number of experts. We use $t\in\{1,\dots,N\}$ for the token index and $e\in\{1,\dots,E\}$ for the expert index. The $t$-th row of $\bm{X}^{(l)}$ is denoted by $\bm{x}^{(l)}_t$.

In a conventional sparse MoE layer, the router computes expert logits solely from the current hidden representations. We denote this original routing pathway by the subscript $O$ and compute its logits as
\begin{equation}
\bm{Z}_{O}^{(l)}
=
\bm{X}^{(l)}
\bm{W}_{O}^{(l)\top}
\in
\mathbb{R}^{N\times E},
\qquad
\bm{W}_{O}^{(l)}
\in
\mathbb{R}^{E\times d},
\end{equation}
where $\bm{W}_{O}^{(l)}$ is the trainable weight block corresponding to the original router.

Given a router logit matrix $\bm{Z}^{(l)}$, where $\bm{Z}^{(l)}=\bm{Z}_{O}^{(l)}$ for a conventional MoE layer, the standard softmax-based sparse MoE dispatch procedure first converts the logit vector of token $t$, denoted by $\bm{Z}^{(l)}_t$, into a probability distribution over the $E$ experts:
\begin{equation}
\bm{Q}^{(l)}_t
=
\operatorname{softmax}
\left(
\bm{Z}^{(l)}_t
\right)
\in
\mathbb{R}^{E}.
\end{equation}
For each token, let $k$ be the number of selected experts. The operator $\operatorname{TopK}(\cdot,k)$ returns the indices of the $k$ largest entries, yielding the selected expert set
\begin{equation}
\mathcal{A}_t
=
\operatorname{TopK}
\left(
\bm{Q}^{(l)}_t,k
\right),
\qquad
|\mathcal{A}_t|=k.
\end{equation}
For a selected expert $e\in\mathcal{A}_t$, using $e'$ as a dummy expert index, the selected probabilities are converted into normalized routing weights as
\begin{equation}
w_{t,e}
=
\frac{
Q^{(l)}_{t,e}
}{
\sum_{e'\in\mathcal{A}_t}
Q^{(l)}_{t,e'}
+\epsilon
},
\qquad
e\in\mathcal{A}_t,
\end{equation}
where $\epsilon>0$ is a small constant for numerical stability. The outputs of the selected expert networks are then aggregated according to these normalized weights:
\begin{equation}
\bm{y}^{(l)}_t
=
\sum_{e\in\mathcal{A}_t}
w_{t,e}
f_e\!\left(
\bm{x}^{(l)}_t
\right),
\end{equation}
where $f_e(\cdot)$ denotes the network of expert $e$ and $\bm{y}^{(l)}_t$ is the output for token $t$.

To incorporate routing history while preserving this standard MoE pathway, we propose the history-aware residual router, referred to as \textsc{HERO-MoE}. Let $M_l$ denote the number of preceding MoE layers whose routing distributions are visible to layer $l$, let $\bm{H}^{(l)}\in\mathbb{R}^{N\times M_lE}$ denote the corresponding concatenated routing memory, and let $\bm{\rho}^{(l)}\in\mathbb{R}^{N\times 1}$ denote a token-wise scale-preserving coefficient vector broadcast along the feature dimension. We use $\odot$ for element-wise multiplication with row-wise broadcasting. The construction of $\bm{H}^{(l)}$ and the computation of $\bm{\rho}^{(l)}$ are detailed in Sections~\ref{subsec:stage_local_routing} and~\ref{subsec:scale_preserving_residual}, respectively. The final router logits are defined as
\begin{equation}
\bm{Z}^{(l)} = \bm{Z}_{O}^{(l)} + \bm{Z}_{R}^{(l)}, \qquad \bm{Z}_{R}^{(l)} = \left(\bm{\rho}^{(l)} \odot \bm{H}^{(l)}\right)\bm{W}_{R}^{(l)\top}, \qquad \bm{W}_{R}^{(l)} \in \mathbb{R}^{E\times M_lE},
\end{equation}
where the subscript $R$ denotes the residual routing branch and $\bm{W}_{R}^{(l)}$ is its trainable weight block. In the implementation, $\bm{X}^{(l)}$ and $\bm{\rho}^{(l)}\odot\bm{H}^{(l)}$ are concatenated along the feature dimension and passed through a single linear projection whose weight is conceptually partitioned as $[\bm{W}_{O}^{(l)}\ \| \bm{W}_{R}^{(l)}]$; the additive form above is algebraically equivalent to this implementation. When no historical layer is available, $M_l=0$ and $\bm{Z}_{R}^{(l)}=\bm{0}$, so $\bm{Z}^{(l)}=\bm{Z}_{O}^{(l)}$. The resulting logits $\bm{Z}^{(l)}$ are then passed to the configured sparse-MoE routing policy; the standard softmax, top-$k$ selection, and weighted aggregation above describe the default path.

\subsection{Initialization and Scaling Strategy}
\label{subsec:initialization_scaling}
\label{subsec:scale_preserving_residual}

The original and residual routing branches receive inputs with different effective dimensions. The original branch takes the current hidden representation of dimension $d$, while the residual branch takes a concatenated routing history whose dimension can grow with the depth of the network. Without an appropriate initialization scale, the residual logits could become increasingly large in deeper layers as more historical routing features are incorporated, leading to a depth-dependent routing scale and overly confident initial assignments.

To keep the initial magnitude of the combined router logits comparable across layers, we use a depth-aware initialization bound
\begin{equation}
b_l
=
\frac{1}{
\sqrt{d+(l-1)E}
}.
\end{equation}
The term $d+(l-1)E$ represents the current hidden dimension together with the maximum historical feature dimension available before layer $l$ in a single-stage setting. In the implementation, the router weight is a single matrix
\begin{equation}
\bm{W}^{(l)}
=
\big[\bm{W}_O^{(l)} \| \bm{W}_R^{(l)}\big]
\in
\mathbb{R}^{E\times(d+M_lE)},
\end{equation}

where the residual block is omitted when $M_l=0$. Its entries are initialized from the same uniform distribution:
\begin{equation}
\bm{W}^{(l)}
\sim
\mathcal{U}(-b_l,b_l),
\end{equation}
where $\mathcal{U}(-b_l,b_l)$ denotes the uniform distribution on $[-b_l,b_l]$. The bound depends on the global layer index $l$ rather than the number of locally visible layers, so that the initialization of a given layer is not changed by the choice of pipeline partition. The first MoE layer has no preceding history and therefore uses only the original router block.

We further align the magnitude of the historical routing memory with that of the current hidden representation. When $M_l>0$, let $\operatorname{RMS}(\cdot)$ denote the root-mean-square magnitude computed independently for each row. We define the token-wise scale-preserving coefficient as
\begin{equation}
\bm{\rho}^{(l)}
=
\frac{
\operatorname{RMS}(\bm{X}^{(l)})
}{
\operatorname{RMS}(\bm{H}^{(l)})+\epsilon
}
\cdot
\sqrt{\frac{l-1}{M_l}},
\label{eq:rou}
\end{equation}
where $\epsilon>0$ is a small constant for numerical stability. The first factor performs dynamic magnitude alignment, while the second factor compensates for the reduction in visible historical layers introduced by pipeline parallelism. For any representation vector $\bm{u}\in\mathbb{R}^{D}$ with $i$-th component $u_i$, the RMS magnitude is
\begin{equation}
\operatorname{RMS}(\bm{u})
=
\sqrt{
\frac{1}{D}
\sum_{i=1}^{D}u_i^2
}.
\end{equation}
In the implementation, the RMS statistics are detached from the gradient graph, so $\bm{\rho}^{(l)}$ acts as a stop-gradient scaling factor. Thus, $\rho^{(l)}$ calibrates the effective energy of the historical input across depths and pipeline partitions, while the initialization strategy keeps the initial router-logit scale comparable.

\subsection{Stage-Local Historical Routing}
\label{subsec:stage_local_routing}

In large-scale distributed training, deep models are typically partitioned across multiple pipeline \citep{huang2019gpipe} stages. Transmitting historical routing features across pipeline stage boundaries would introduce additional communication overhead and may disrupt the execution schedule. Therefore, \textsc{HERO-MoE} restricts historical routing retrieval to the current pipeline stage. For layer $l$, we define the set of visible predecessor layers as
\begin{equation}
\mathcal{S}(l)
=
\{\,j \mid j<l,\;
\text{layer }j\text{ resides on the same pipeline stage as layer }l\,\}.
\end{equation}
The number of visible historical layers is
\begin{equation}
M_l=|\mathcal{S}(l)|.
\end{equation}
Thus, only preceding layers on the same pipeline stage contribute to the historical routing memory, avoiding cross-stage communication while preserving the execution schedule.

For each historical layer $j\in\mathcal{S}(l)$, we retrieve its dense, pre-Top-$k$ routing probability distribution:
\begin{equation}
\bm{Q}^{(j)}
=
\operatorname{StopGrad}
\left(
\operatorname{softmax}
\left(
\bm{Z}^{(j)}
\right)
\right)
\in
\mathbb{R}^{N\times E}.
\end{equation}
Here, $\bm{Z}^{(j)}$ denotes the router logits computed from the current representation at layer $j$ and its available historical routing input. The distribution $\bm{Q}^{(j)}$ is stored immediately after these logits are computed and before the subsequent configured top-$k$ routing policy is applied. The operator $\operatorname{StopGrad}(\cdot)$ detaches its input from the gradient graph, preventing gradients from the current router from propagating through previous routing decisions.

Enumerate the visible predecessor indices as $j_1,\ldots,j_{M_l}$, where $j_m$ is the $m$-th visible predecessor for $m\in\{1,\ldots,M_l\}$. The historical routing memory is constructed by concatenating their routing distributions along the feature dimension:
\begin{equation}
\bm{H}^{(l)}
=
\left[
\bm{Q}^{(j_1)},
\bm{Q}^{(j_2)},
\ldots,
\bm{Q}^{(j_{M_l})}
\right]
\in
\mathbb{R}^{N\times M_lE},
\end{equation}
where $\{j_1,\ldots,j_{M_l}\}=\mathcal{S}(l)$. The $t$-th row of $\bm{H}^{(l)}$ corresponds to the routing history of the same token represented by the $t$-th row of $\bm{X}^{(l)}$. At layer $l$, this history is scaled and mapped to $\bm{Z}_{R}^{(l)}$, added to $\bm{Z}_{O}^{(l)}$, and then passed through the standard sparse MoE routing procedure described in Section~\ref{sec:method}.

\section{Experiments}
\label{experiments}

In this section, we conduct empirical evaluation of HERO-MoE. We first detail the experimental setup including datasets, model configurations and evaluation benchmarks, then present the effectiveness results and further analyze the learned inter-layer routing patterns. All training runs were performed on NPUs, each with 64 GB of NPU memory.

\subsection{Experiment Settings}
\noindent\textbf{Datasets}.
% 训练数据集包含开源数据集：dclm \cite {li2024datacomp}、map-cc \citep {du2024chinese}、ultrachat \citep {ding2023enhancing}、tuluv3 \citep {lambert2024tulu}、finemath \citep {liu2024finemath}、megamath \citep {zhou2025megamath}，以及未公开的高质量教育相关专有自行收集数据集。
The training dataset includes open-source datasets: dclm \citep{li2024datacomp}, map-cc \citep{du2024chinese}, ultrachat \citep{ding2023enhancing}, tuluv3 \citep{lambert2024tulu}, finemath \citep{liu2024finemath}, megamath \citep{zhou2025megamath}, as well as high-quality proprietary self-collected datasets related to education that are not publicly available.

\noindent\textbf{Model Configuration}.
%实验部分包含Deepseek和Qwen架构的MOE，默认Deepseek。训练在长度为4k个token 的序列上进行，总共训练约100B token。
The experimental evaluation defaults to a MoE model built with our custom architecture (close to Deepseek-MoE \citep{liu2024deepseekv3}), which has approximately 8B total parameters and 0.5B activated parameters. Pre-training with random initialization adopts sequences of 4K tokens, with a total training corpus of roughly 100B tokens. Specifically, the model comprises 30 Transformer layers with a hidden dimension of 1024. The attention layer adopts Multi-Head Latent Attention \citep{liu2024deepseek} (MLA, 16 attention heads, Query LoRA rank of 512). The feed-forward layer employs an MoE architecture with 128 experts (intermediate size of 640), dynamically routing to Top-4 experts per token. The learning rate follows a cosine decay schedule from $2\times 10^{-3}$ to $5\times 10^{-5}$ with a global batch size of 1024 across 25,000 training steps. 
The training results before introducing HERO-MoE are denoted as Baseline.
% 实验评估默认采用基于我们自定义架构构建的 MoE 模型，该架构接近 DeepSeek-MoE~\citep{liu2024deepseekv3}，模型总参数量约为 8B，激活参数量约为 0.5B。模型从随机初始化开始进行预训练，采用 4K token 的序列长度，总训练语料规模约为 100B tokens。具体而言，该模型包含 30 个 Transformer 层，隐藏维度为 1024。注意力层采用 Multi-Head Latent Attention~\citep{liu2024deepseek}（MLA，16 个注意力头，Query LoRA rank 为 512）。前馈层采用 MoE 架构，包含 128 个专家（中间层维度为 640），并为每个 token 动态路由到 Top-4 个专家。学习率采用 cosine decay 策略，在 25,000 个训练 step 内从 $2\times 10^{-3}$ 衰减到 $5\times 10^{-5}$，global batch size 为 1024。加入HERO-MoE前的训练结果由Baseline表示。

\noindent\textbf{Evaluation Metrics}.
The general benchmarks cover 13 typical task categories, including mathematical reasoning (GSM8K \citep{cobbe2021training}, MathQA \citep{amini2019mathqa}, MATH \citep{hendrycks2021measuring}), commonsense question answering (ARC-C, ARC-C-ZH, ARC-E \citep{clark2018think}), comprehensive cognitive reasoning (MMLU \citep{hendrycks2020measuring}, CMMLU \citep{li2024cmmlu}, BBH \citep{suzgun2023challenging}), text coherence assessment (LAMBADA \citep{paperno2016lambada}), code generation (HumanEval \citep{chen2021evaluating}), and domain-specific tasks (HellaSwag \citep{zellers2019hellaswag}, PiPA \citep{kim2025pipa}), which span both English and Chinese bilingual scenarios.

\subsection{Effectiveness of HERO-MoE}

\begin{figure*}[t]
  \includegraphics[width=\linewidth]{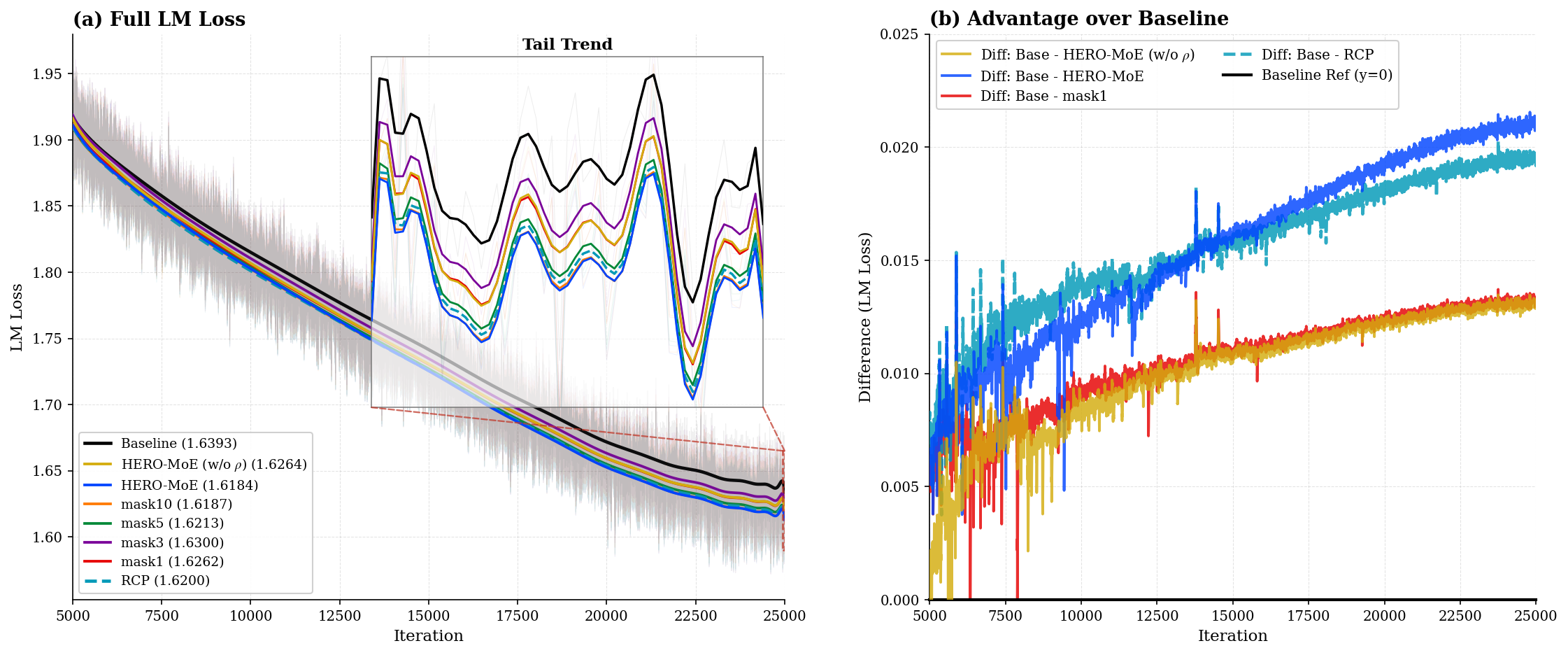}
  \caption{Training loss of the Baseline, RCP, HERO-MoE, historical-window ablations, and the no-scaling ablation.}
  \label{fig:loss_comparison}
\end{figure*}

\noindent\textbf{Training Loss}.
Figure~\ref{fig:loss_comparison}(a) presents the full training-loss trajectories of different routing-enhancement methods, with the inset providing a zoomed-in view of the final training stage. Here, the baseline denotes the raw MoE model without historical routing, RCP~\citep{hu2026synergistic} denotes the adjacent-layer route consolidation method, and mask1, mask3, mask5, and mask10 denote ablation variants that use only recent historical layers. The no-scaling ablation removes the scale-preserving factor in Eq.~(\ref{eq:rou}) while retaining the historical routing branch. The historical-routing variants generally achieve lower final losses than the baseline. Full HERO-MoE achieves the lowest final loss, around 1.6184 compared with 1.6393 for the baseline, while mask10 performs very closely, indicating that a medium-range history captures most of the benefit. The no-scaling variant also benefits from historical routing, but converges to a higher final loss than the full method, showing that the scale-preserving operation is important for effectively incorporating routing history. RCP also improves over the baseline, but remains weaker than HERO-MoE and mask10, showing that, in our setting, HERO-MoE achieves lower training loss without introducing an additional loss objective or requiring hyperparameter tuning.
% 图~\ref{fig:loss_comparison}(a) 展示了不同路由增强方法的完整训练损失曲线，插图进一步放大了训练末期的变化趋势。其中，Baseline 表示不使用历史路由的原始 MoE 模型，RCP~\citep{hu2026synergistic} 表示相邻层路径巩固方法，mask1、mask3、mask5 和 mask10 表示仅使用最近若干层历史信息的消融设置。去除缩放操作的消融方法在保留历史路由分支的同时，移除了公式~\ref{eq:rou} 中的尺度保持因子。总体来看，历史路由增强方法通常能够取得低于 Baseline 的最终损失。完整 HERO-MoE 达到最低的最终损失，约为 1.6184，而 Baseline 为 1.6393；mask10 的表现与完整方法非常接近，说明中等范围的历史信息已经能够捕获大部分收益。去除缩放操作的模型同样能够从历史路由中获得收益，但其最终损失高于完整 HERO-MoE，说明尺度保持操作对于有效利用历史路由信息具有重要作用。RCP 同样优于 Baseline，但仍弱于 HERO-MoE 和 mask10，说明在当前实验设置下，HERO-MoE 无需引入额外的损失函数或进行方法特定调参，也能取得更低的训练损失。

Figure~\ref{fig:loss_comparison}(b) reports the loss advantage over the baseline throughout training. The full HERO-MoE method maintains the largest advantage among the compared variants, whereas the no-scaling ablation exhibits a smaller improvement, further demonstrating the value of the scale-preserving operation. The ablation results also show that an overly short history limits the performance gain: mask1 provides a noticeably smaller improvement, suggesting that routing decisions benefit from richer historical trajectories beyond the immediately preceding layer. Moreover, the advantage of HERO-MoE over the baseline continues to increase during training, whereas the improvement of RCP saturates earlier. These results indicate that HERO-MoE provides sustained benefits for long-term MoE optimization.
% 图~\ref{fig:loss_comparison}(b) 展示了整个训练过程中各方法相对于 Baseline 的损失优势。完整 HERO-MoE 在所比较的方法中保持了最大的优势，而去除缩放操作的消融方法带来的提升较小，进一步说明了尺度保持操作的有效性。消融结果还表明，过短的历史窗口会限制性能增益：mask1 的提升明显较小，说明路由决策能够从超越紧邻前一层的、更丰富的历史轨迹中获益。此外，HERO-MoE 相对于 Baseline 的优势在训练过程中持续扩大，而 RCP 的提升则较早趋于饱和。这些结果表明，HERO-MoE 能够在长期 MoE 优化过程中持续带来收益。

\begin{table*}[t]
\tiny
\setlength{\tabcolsep}{2pt}
\caption{General benchmark comparison between the baseline MoE and HERO-MoE.}
\centering
\begin{tabular}{l|ccccccccccccc|c}
\toprule
\textbf{Model} & \textbf{GSM8K} & \textbf{MathQA} & \textbf{MATH} & \textbf{ARC-C} & \textbf{ARC-C-ZH} & \textbf{ARC-E} & \textbf{HellaSwag} & \textbf{BBH} & \textbf{MMLU} & \textbf{CMMLU} & \textbf{HumanEval} & \textbf{LAMBADA} & \textbf{PIPA} & \textbf{Avg.} \\
\midrule
\textbf{Baseline}    & 10.99 & \textbf{30.99} & 8.14 & 35.32 & 30.97 & \textbf{70.96} & 40.62 & \textbf{29.3} & 42.1 & \textbf{43.2} & 15.24 & 45.55 & 72.03 & 36.57 \\
\textbf{HERO-MoE} & \textbf{12.21} & 30.95 & \textbf{9.6} & \textbf{37.63} & \textbf{31.83} & 70.92 & \textbf{42.01} & 29.09 & \textbf{42.22} & 42.38 & \textbf{18.9} & \textbf{46.52} & \textbf{72.2} & \textbf{37.42} \\
\midrule
\textbf{$\Delta$} 
& \textcolor{green!70!black}{1.22} 
& \textcolor{red!70!black}{$-$0.04} 
& \textcolor{green!70!black}{1.46} 
& \textcolor{green!70!black}{2.31} 
& \textcolor{green!70!black}{0.86} 
& \textcolor{red!70!black}{$-$0.04} 
& \textcolor{green!70!black}{1.39} 
& \textcolor{red!70!black}{$-$0.21} 
& \textcolor{green!70!black}{0.12} 
& \textcolor{red!70!black}{$-$0.82} 
& \textcolor{green!70!black}{3.66} 
& \textcolor{green!70!black}{0.97} 
& \textcolor{green!70!black}{0.17} 
& \textcolor{green!70!black}{0.85} \\
\bottomrule
\end{tabular}
\label{tab:base_hero_bench}
\end{table*}

\noindent\textbf{General Benchmarks}.
Table~\ref{tab:base_hero_bench} compares the baseline MoE and HERO-MoE on multiple general benchmarks. Overall, our method improves the average score from 36.57 to 37.42, bringing an average gain of +0.85. Among the 13 tasks, HERO-MoE achieves improvements on 9 tasks.

The most notable gains appear on HumanEval (+3.66), ARC-C (+2.31), MATH (+1.46), HellaSwag (+1.39), and GSM8K (+1.22). These tasks usually require stronger reasoning, code generation, or complex semantic modeling abilities, indicating that HERO-MoE improves not only the training loss but also downstream task performance. Meanwhile, HERO-MoE shows slight drops on MathQA, ARC-E, BBH, and CMMLU, suggesting that historical routing enhancement is not monotonically beneficial for all tasks. One possible reason is that historical routing changes the expert specialization pattern, making its benefit less obvious on some knowledge-oriented or multiple-choice tasks.

Overall, by introducing historical routing information, HERO-MoE improves both the training behavior and generalization performance of MoE models. The average performance gain is stable, with especially clear benefits on mathematics, coding, and complex reasoning tasks.

% 表格比较了 Base 和 Quanshi/HERO-MoE 在多个通用 benchmark 上的结果。整体来看，我们的方法将平均分从 36.57 提升到 37.42，带来 +0.85 的平均增益。在 13 个任务中，HERO 在 9 个任务上取得提升。

% 提升较明显的任务包括 HumanEval（+3.66）、ARC-C（+2.31）、MATH（+1.46）、HellaSwag（+1.39）和 GSM8K（+1.22）。这些任务通常需要更强的推理、代码生成或复杂语义建模能力，说明 HERO 改善的不只是训练 loss，也能转化为下游能力提升。与此同时，MathQA、ARC-E、BBH 和 CMMLU 上存在小幅下降，说明历史路由增强并非对所有任务都单调有利，可能会改变专家分工，使部分知识型或选择题任务的收益不如推理类任务明显。

% 总体而言，HERO-MoE通过引入历史路由信息改善了 MoE 的训练与泛化表现，平均性能提升稳定，尤其在数学、代码和复杂推理任务上收益更明显。

\subsection{Findings of HERO-MoE}

\begin{figure*}[t]
\centering
  \includegraphics[width=\linewidth]{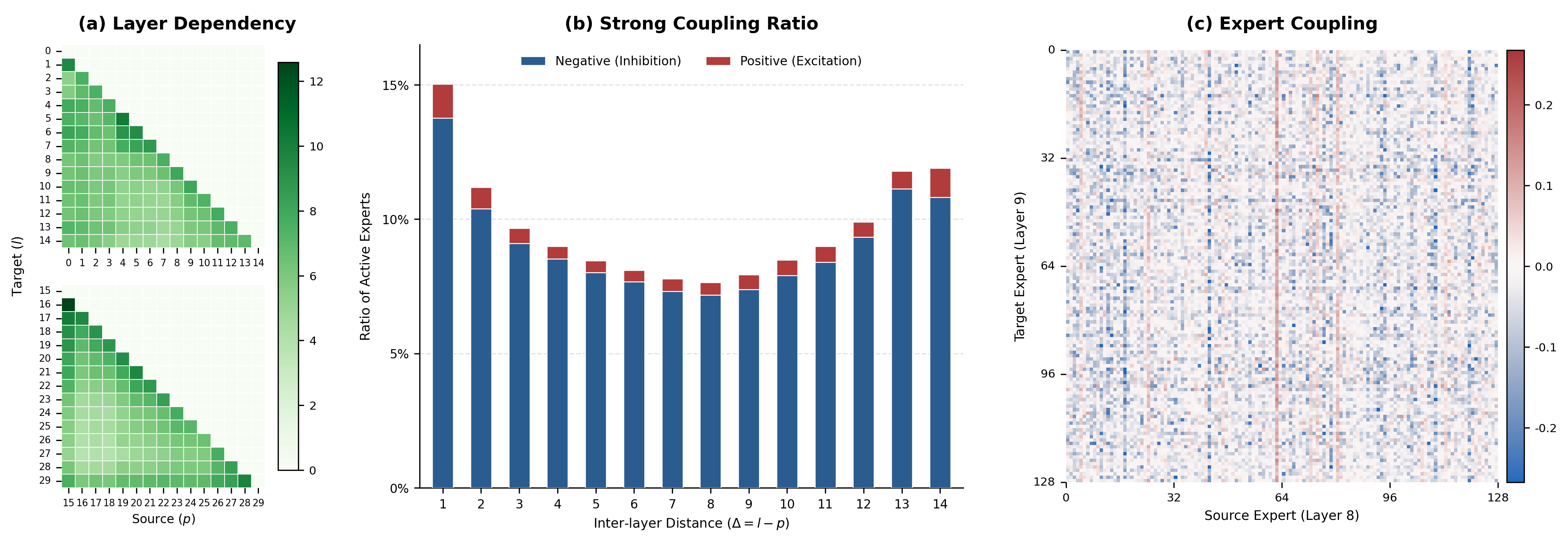}
  \caption{Layer-wise historical dependencies, distance-wise strong coupling ratios, and a representative expert-level coupling matrix learned by HERO-MoE.}
  \label{fig:routing_analysis}
\end{figure*}

Figure~\ref{fig:routing_analysis}(a) visualizes the layer-wise dependency strength learned by the historical routing branch in HERO-MoE. Each lower-triangular entry represents the routing dependency from a predecessor layer $p$ to a target layer $l$, where darker green indicates a stronger influence of the historical layer on the current router. We observe that the learned dependencies are not limited to adjacent layers, but extend across multiple preceding layers. Although entries near the diagonal are generally stronger, visible non-zero dependencies also appear between more distant layers. This result suggests that routing history forms a cross-depth trajectory rather than a purely adjacent-layer effect, providing useful routing priors for subsequent MoE layers.

To quantify how expert-level interactions vary with layer distance, Figure~\ref{fig:routing_analysis}(b) reports the fraction of residual-router weight entries exhibiting strong negative or positive coupling at each inter-layer distances $\Delta=l-p$. Under this fixed global threshold, adjacent-layer pairs have the largest fraction of strongly coupled weight entries, while this fraction generally decreases over intermediate distances. However, the ratio rises again for larger distances, especially at $\Delta=13$ and $\Delta=14$, indicating that while short-range interactions dominate overall, a non-negligible subset of distant layer pairs still exhibits strong expert-level dependencies. Across distances, negative couplings account for the larger proportion, suggesting that cross-layer routing history often encodes expert competition or suppression in addition to cooperative relationships.

Figure~\ref{fig:routing_analysis}(c) provides a representative expert-level coupling matrix from Layer 8 to Layer 9. The horizontal axis denotes source-layer experts, and the vertical axis denotes target-layer experts. Each entry measures the influence of a source expert on a target expert: red indicates positive cooperation, blue indicates negative competition, and white indicates a near-neutral relation. Vertical red and blue patterns indicate source experts that broadly promote or suppress multiple target experts, whereas vertical white patterns indicate source experts with weak influence on most subsequent experts. Horizontal white patterns indicate target experts that are relatively insensitive to historical expert choices and may be determined primarily by the current token representation. Scattered dark red and dark blue entries further reveal strong pairwise cooperative and competitive relations between specific expert pairs. Additional expert-level coupling matrices for other layer pairs are provided in Appendix Figure~\ref{fig:all_expert}.

Overall, Figure~\ref{fig:routing_analysis} provides a hierarchical view of the routing structures learned by HERO-MoE and offers a direct perspective for studying cross-layer routing dynamics, expert specialization, and expert composition mechanisms in sparse MoE models.

\subsection{Cost of HERO-MoE}

\begin{table*}[t]
\centering
\tiny
\caption{Time, memory and FLOPs comparison between Baseline and HERO-MoE.}
\label{tab:quanshi_efficiency}
\begin{tabular}{l|cccccc}
\toprule
\textbf{Method} & \textbf{Peak Memory} & \textbf{FLOPs (Router only)} & \textbf{FLOPs (Total)} & \textbf{Forward Time} & \textbf{Router Step Time} & \textbf{Total Train Time} \\
\midrule
Baseline
& 5440.8\,MiB
& 1.55\,T
& 211.83\,T
& 32.579\,ms
& 0.14\,s / iter
& 71.88\,h \\
HERO-MoE
& 5464.9\,MiB
& 2.90\,T
& 213.18\,T
& 37.984\,ms
& 0.54\,s / iter
& 74.65\,h \\
\midrule
$\Delta$
& +0.44\%
& +87.50\%
& +0.64\%
& +16.59\%
& +281.14\%
& +3.85\% \\
\bottomrule
\end{tabular}
\end{table*}

As shown in Table~\ref{tab:quanshi_efficiency}, we evaluate the computational cost of HERO-MoE against the Baseline. HERO-MoE introduces only a small memory overhead, with peak training memory increasing by 0.44\% (+24.2\,MiB). The expanded gating dimension increases the standalone router FLOPs from 1.55\,T to 2.90\,T (+87.50\%) and the router step time from 0.14\,s/iter to 0.54\,s/iter. However, because routing accounts for only a small fraction of the full model computation, the total model step FLOPs increase from 211.83\,T to 213.18\,T (+0.64\%). The end-to-end forward latency increases by 5.405\,ms ($\approx 0.18$\,ms per layer), and the overall training wall-clock time increases from 71.88\,h to 74.65\,h over 25,000 steps, corresponding to a 3.85\% overhead. These measurements show that HERO-MoE introduces noticeable router-side overhead but maintains modest end-to-end training overhead in our evaluated setting.

\section{Conclusion}
\label{sec:conclusion}

In this paper, we introduced HERO-MoE, a historical expert routing framework for sparse Mixture-of-Experts models. Instead of making routing decisions solely from the current token representation, HERO-MoE reuses detached routing distributions from preceding MoE layers within the same pipeline stage and injects them into the current router through a scale-preserving historical residual branch. This design keeps the standard sparse dispatch procedure unchanged, avoids cross-stage communication, and introduces historical routing priors without adding an auxiliary routing loss.

% 本文提出了 HERO-MoE，一个面向稀疏 Mixture-of-Experts 模型的历史专家路由框架。不同于仅基于当前 token 表征进行路由决策的方法，HERO-MoE 复用同一 pipeline stage 内前序 MoE 层产生的、已分离梯度的路由分布，并通过保尺度的历史残差分支将其注入当前 router。该设计保持标准稀疏派发流程不变，避免跨 stage 通信，并且在不引入额外路由辅助损失的情况下加入历史路由先验。

Experiments show that HERO-MoE improves both training dynamics and downstream performance. Compared with the baseline MoE, HERO-MoE reduces the final training loss from 1.6393 to 1.6184 and improves the average benchmark score from 36.57 to 37.42 across 13 evaluation tasks. Ablation studies further show that using richer historical routing trajectories is more effective than relying only on very short history windows. In addition, the trained historical router provides a direct way to visualize layer-wise routing dependencies and expert-level cooperative or competitive relations, offering new insights into how sparse expert models organize routing behavior across depth.

% 实验结果表明，HERO-MoE 同时改善了训练动态和下游性能。相比 baseline MoE，HERO-MoE 将最终训练 loss 从 1.6393 降低到 1.6184，并在 13 个通用评测任务上将平均分从 36.57 提升到 37.42。消融实验进一步表明，使用更丰富的历史路由轨迹比仅依赖很短的历史窗口更加有效。此外，训练得到的历史 router 提供了一种直接可视化层间路由依赖和专家级协同/互斥关系的方式，为理解稀疏专家模型如何在深度方向上组织路由行为提供了新的视角。

Overall, our results suggest that routing history is a useful and underexplored signal for improving MoE training. By incorporating historical routing information in a lightweight and stage-local manner, HERO-MoE provides a simple, practical, and interpretable direction for building more effective sparse expert models. A current limitation is that routing information is not shared across pipeline-parallel stages, which may limit HERO-MoE's effectiveness at high pipeline-parallel degrees. Future work will extend HERO-MoE to support cross-stage information transfer.

% 总体而言，HERO-MoE 表明路由历史是一个有用但尚未被充分探索的信号，可用于改善 MoE 训练。通过以轻量级、stage-local 的方式引入历史路由信息，HERO-MoE 为构建更有效的稀疏专家模型提供了一个简单、实用且可解释的方向。当前的一个局限是，我们需要在后续工作中尽快支持跨pp信息传递以提升HERO-MoE 在高pp场景下的有效性。

\section*{AI Assistance}
The AI support mainly covers text fluency optimization, logical structure sorting and grammar correction, and all related outputs have been manually verified and revised by the authors.

\bibliography{iclr2027_conference}
\bibliographystyle{iclr2027_conference}

\newpage
\appendix
\section*{APPENDIX}
\section{More Expert-level Samples}

\begin{figure*}[t]
  \includegraphics[width=\linewidth]{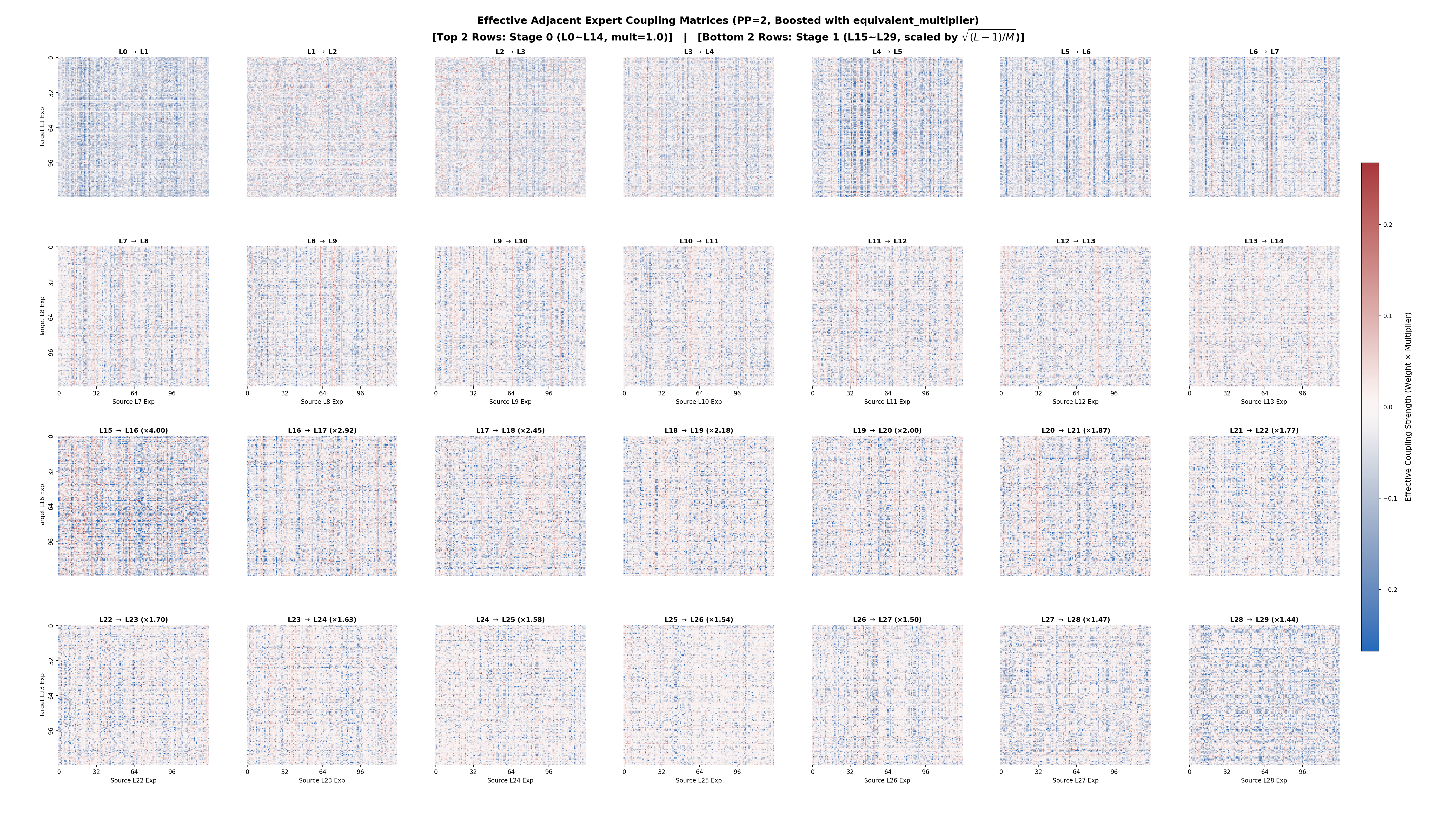}
  \caption{Expert-level coupling matrices for all adjacent MoE-layer pairs learned by HERO-MoE.}
  \label{fig:all_expert}
\end{figure*}

To complement the representative coupling matrix shown in Figure \ref{fig:routing_analysis}(c), Figure 4 presents the expert-level coupling matrices for all adjacent MoE-layer pairs, and the later fourteen plots are multiplied by the second factor in Eq.~(\ref{eq:rou}). Across different layer pairs, the matrices exhibit structured positive and negative regions rather than unstructured noise, indicating that the historical routing branch captures non-trivial cooperation and competition between experts. These patterns are not identical across depth: some source experts exert broad promoting or suppressing effects on multiple target experts, whereas others have only localized or weak influence. Likewise, several target experts remain relatively insensitive to the preceding routing state, suggesting that their assignments are determined primarily by the current token representation. Overall, Figure 4 shows that the coupling observed in Figure \ref{fig:routing_analysis}(c) is not an isolated case, but reflects a broader, layer-dependent organization of cross-layer expert interactions learned by HERO-MoE.

\section{Expert Routing Dynamics}
\label{sec:expert_routing}

\begin{figure*}[t]
  \centering
  \includegraphics[width=\linewidth]{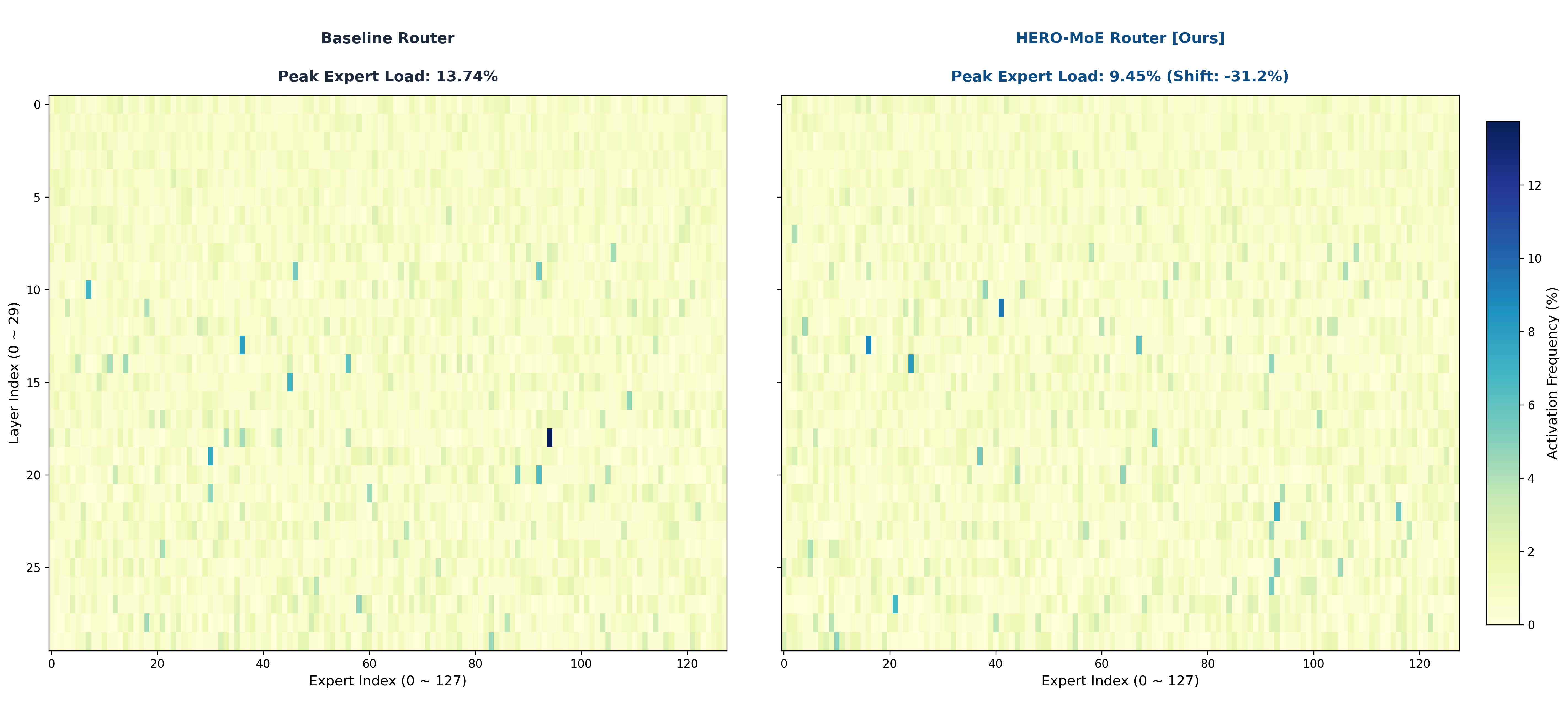}
  \caption{Expert utilization frequency across 30 layers and 128 experts.}
  \label{fig:expert_frequency}
\end{figure*}

Figure~\ref{fig:expert_frequency} illustrates the layer-wise expert activation distribution across all 30 layers and 128 experts, evaluated on multiple benchmarks. In the baseline model, Expert 94 in Layer 18 monopolizes 13.74\% of all tokens, creating a severe bottleneck and exacerbating the straggler problem in distributed All-to-All communication. In contrast, HERO-MoE reduces the peak expert load by 31.2\%, down to 9.45\%, thereby effectively alleviating cluster communication congestion. More importantly, HERO-MoE accelerates model training convergence while maintaining balanced expert utilization.

\section{Additional Results on Qwen3}

\begin{figure*}[t]
  \centering
  \includegraphics[width=\linewidth]{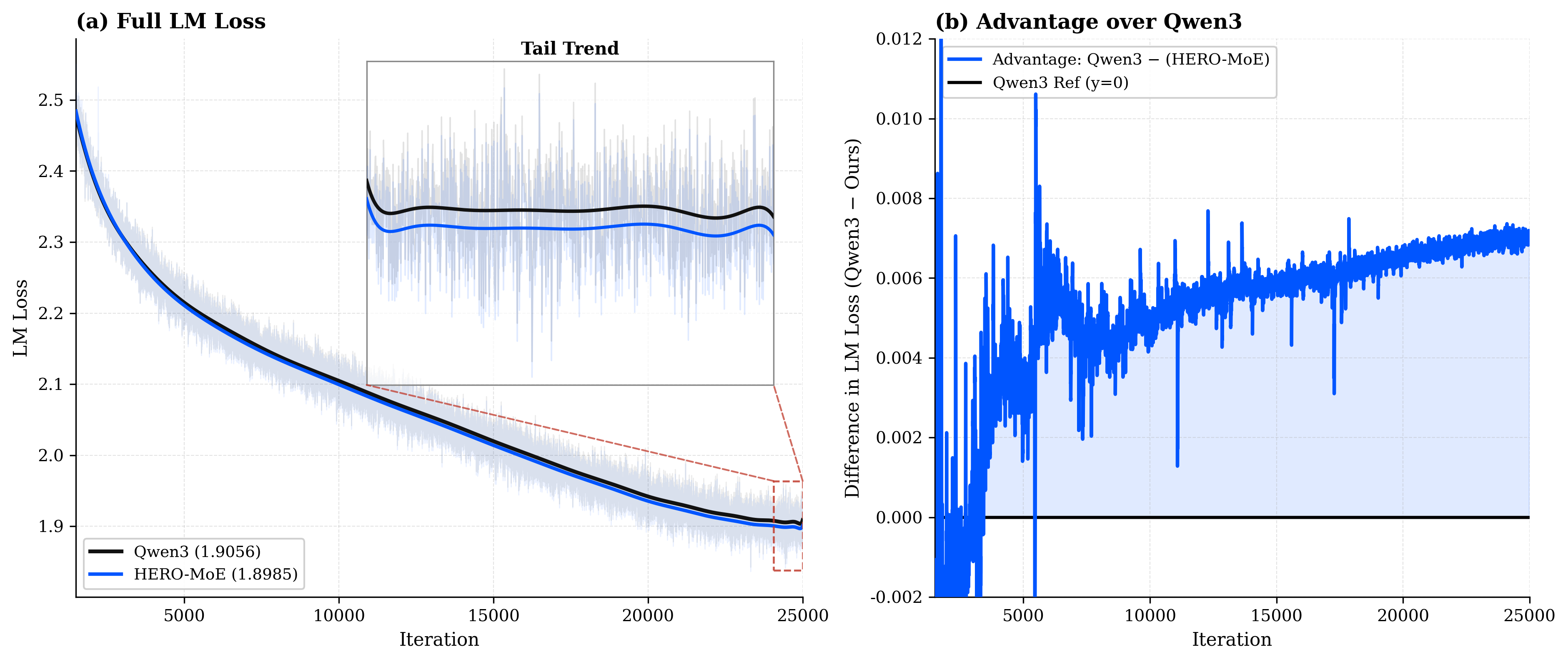}
  \caption{Training loss and loss advantage of Qwen3 with HERO-MoE.}
  \label{fig:lossqwen}
\end{figure*}

Figure~\ref{fig:lossqwen}(a) compares the training loss of Qwen3 \citep{yang2025qwen3} with and without HERO-MoE under the Qwen3 experimental configuration. HERO-MoE generally achieves a lower loss, especially during the later training stage, reducing the final loss from 1.9056 to 1.8985. Figure~\ref{fig:lossqwen}(b) reports the loss advantage over the original Qwen3 model, computed as the Qwen3 loss minus the loss of Qwen3 with HERO-MoE. Without model-specific tuning, although the advantage fluctuates considerably during early training, it remains predominantly positive and continues to increase during the later stage. These results further demonstrate the generality of HERO-MoE across MoE models with different architectures.
% 图~\ref{fig:lossqwen}(a)比较了在Qwen3实验配置下，使用和不使用HERO-MoE时的训练损失。HERO-MoE通常能够取得更低的损失，尤其是在训练后期，最终损失从1.9056降至1.8985。图~\ref{fig:lossqwen}(b)展示了相对于原始Qwen3模型的损失优势，其计算方式为Qwen3的损失减去结合HERO-MoE后的损失。未经特定调优的情况下，尽管训练初期的优势存在较大波动，但在后期大部分时间保持为正，并稳定增长。结果进一步表明，HERO-MoE对于不同架构的MoE模型的普适性。

\section{Mathematical Properties of HERO-MoE}
\label{app:mathematical_properties}
This appendix formalizes the structural properties of HERO-MoE. We use
the same notation as in the method section. In particular, let
$\mathcal{S}(l)$ be the set of preceding MoE layers on the same pipeline
stage as layer $l$, and let $M_l=|\mathcal{S}(l)|$. The results below concern the explicit historical
routing path. They establish structural and optimization properties of
the proposed parameterization; improvements in training loss, expert
utilization, and downstream accuracy remain empirical claims.

\subsection{Notation and assumptions}
For $M_l>0$, write the detached historical distributions as
\begin{equation}
\bm{Q}^{(j)}
=
\operatorname{StopGrad}\!\left(
\operatorname{softmax}\!\left(\bm{Z}^{(j)}\right)
\right)
\in\mathbb{R}^{N\times E},
\qquad j\in\mathcal{S}(l),
\end{equation}
and define
\begin{equation}
\bm{H}^{(l)}
=
\left[
\bm{Q}^{(j_1)},
\bm{Q}^{(j_2)},
\ldots,
\bm{Q}^{(j_{M_l})}
\right]
\in\mathbb{R}^{N\times M_lE}.
\end{equation}

For a token row, let
\begin{equation}
r_x
=
\operatorname{RMS}\!\left(\bm{x}^{(l)}_t\right),
\qquad
r_h
=
\operatorname{RMS}\!\left(\bm{h}^{(l)}_t\right).
\end{equation}
Define the scale-preserving coefficient as
\begin{equation}
\rho_t^{(l)}
=
\frac{r_x}{r_h+\epsilon}
\sqrt{\frac{l-1}{M_l}},
\qquad
\epsilon>0.
\label{eq:app_rho}
\end{equation}
When an exact identity is stated, we first consider the idealized limit
$\epsilon=0$ and then give the corresponding expression for the
finite-$\epsilon$ case.

\subsection{History-aware router decomposition}
\paragraph{Proposition 1 (Baseline recovery).}
If $M_l=0$, HERO-MoE exactly reduces to the conventional linear MoE
router. If $M_l>0$, the history-aware router is algebraically equivalent
to an original-state branch plus a historical residual branch.

\emph{Proof.}
Partition the router weight as
\begin{equation}
\bm{W}^{(l)}
=
\left[
\bm{W}_{O}^{(l)}
\middle\|
\bm{W}_{R}^{(l)}
\right],
\qquad
\bm{W}_{O}^{(l)}\in\mathbb{R}^{E\times d},
\quad
\bm{W}_{R}^{(l)}\in\mathbb{R}^{E\times M_lE}.
\end{equation}
The input to the single linear projection is
\begin{equation}
\bm{U}^{(l)}
=
\left[
\bm{X}^{(l)}
\middle\|
\bm{\rho}^{(l)}\odot\bm{H}^{(l)}
\right].
\end{equation}
Block matrix multiplication gives
\begin{equation}
\begin{aligned}
\bm{Z}^{(l)}
&=
\bm{U}^{(l)}\bm{W}^{(l)\top} \\
&=
\bm{X}^{(l)}\bm{W}_{O}^{(l)\top}
+
\left(
\bm{\rho}^{(l)}\odot\bm{H}^{(l)}
\right)
\bm{W}_{R}^{(l)\top}.
\end{aligned}
\end{equation}
When $M_l=0$, the second block is absent, so
$\bm{Z}^{(l)}=\bm{X}^{(l)}\bm{W}_{O}^{(l)\top}$, which is the baseline
router. \hfill$\square$

\paragraph{Corollary 1 (Unchanged sparse dispatch).}
HERO-MoE changes only the logits supplied to the configured routing
operator. Consequently, the subsequent softmax, top-$k$ selection,
group-limited selection, capacity handling, and token dispatch operations
are unchanged. In particular, for any routing operator
$\mathcal{R}$ acting on router logits,
\begin{equation}
\operatorname{HERO\text{-}MoE\ routing}
=
\mathcal{R}\!\left(
\bm{Z}_{O}^{(l)}+\bm{Z}_{R}^{(l)}
\right),
\end{equation}
with the same operator $\mathcal{R}$ as in the baseline model.

\subsection{Gradient isolation of historical routing}
\paragraph{Proposition 2 (Stop-gradient isolation).}
Let $\theta_j$ denote parameters that produce the router logits
$\bm{Z}^{(j)}$ of a preceding layer. Along the explicit historical path,
the current layer satisfies
\begin{equation}
\frac{\partial \bm{H}^{(l)}}{\partial\theta_j}
=
\bm{0},
\qquad
\frac{\partial \bm{Z}_{R}^{(l)}}{\partial\theta_j}
=
\bm{0}.
\label{eq:app_stopgrad}
\end{equation}

\emph{Proof.}
For every visible predecessor $j$, the historical routing distribution is
\begin{equation}
\bm{Q}^{(j)}
=
\operatorname{StopGrad}\!\left(
\operatorname{softmax}\!\left(\bm{Z}^{(j)}\right)
\right).
\end{equation}
By definition of the stop-gradient operator, its value is preserved in
the forward pass while its derivative with respect to its input is zero.
Therefore,
\begin{equation}
\frac{\partial\bm{Q}^{(j)}}{\partial\theta_j}
=
\bm{0}.
\end{equation}
Concatenation is linear with respect to its inputs, so the same property
holds for $\bm{H}^{(l)}$. Since the RMS statistics used in
$\bm{\rho}^{(l)}$ are also treated as stop-gradient quantities, the product
$\bm{\rho}^{(l)}\odot\bm{H}^{(l)}$ has no gradient through the explicit
historical path. Multiplication by the trainable matrix
$\bm{W}_{R}^{(l)\top}$ therefore yields Eq.~(\ref{eq:app_stopgrad}).
This statement does not remove gradients that may reach earlier layers
through the ordinary hidden-state path $\bm{X}^{(l)}$. \hfill$\square$

\subsection{Scale-preserving historical fusion}
\paragraph{Proposition 3 (RMS calibration identity).}
For a token with $r_h>0$, the scaled historical vector
\begin{equation}
\widetilde{\bm{h}}^{(l)}_t
=
\rho_t^{(l)}\bm{h}^{(l)}_t
\end{equation}
satisfies
\begin{equation}
\operatorname{RMS}\!\left(
\widetilde{\bm{h}}^{(l)}_t
\right)
=
r_x
\sqrt{\frac{l-1}{M_l}}
\frac{r_h}{r_h+\epsilon}.
\label{eq:app_rms}
\end{equation}
In the idealized case $\epsilon=0$,
\begin{equation}
\operatorname{RMS}\!\left(
\widetilde{\bm{h}}^{(l)}_t
\right)
=
r_x\sqrt{\frac{l-1}{M_l}}.
\label{eq:app_rms_exact}
\end{equation}

\emph{Proof.}
The coefficient $\rho_t^{(l)}$ is a nonnegative scalar shared by all
components of the token row. Hence
\begin{equation}
\operatorname{RMS}\!\left(
\rho_t^{(l)}\bm{h}^{(l)}_t
\right)
=
\rho_t^{(l)}
\operatorname{RMS}\!\left(
\bm{h}^{(l)}_t
\right)
=
\rho_t^{(l)}r_h.
\end{equation}
Substituting Eq.~(\ref{eq:app_rho}) proves Eq.~(\ref{eq:app_rms}),
and setting $\epsilon=0$ proves Eq.~(\ref{eq:app_rms_exact}).
\hfill$\square$

\paragraph{Corollary 2 (Equivalent historical energy).}
Let the historical vector have dimension $M_lE$. For $\epsilon=0$,
\begin{equation}
\left\|
\widetilde{\bm{h}}^{(l)}_t
\right\|_2^2
=
(l-1) E r_x^2.
\label{eq:app_energy_exact}
\end{equation}
Thus, after RMS alignment, the visible history has the same squared
energy as $l-1$ virtual historical blocks of width $E$, regardless of the
number $M_l$ of blocks that are locally visible. For $\epsilon>0$, the
finite-$\epsilon$ expression is
\begin{equation}
\left\|
\widetilde{\bm{h}}^{(l)}_t
\right\|_2^2
=
(l-1) E r_x^2
\left(
\frac{r_h}{r_h+\epsilon}
\right)^2,
\label{eq:app_energy_eps}
\end{equation}
which differs from the ideal identity only by the numerical-stability
factor in parentheses. If $r_h=0$, the historical vector is already zero;
the positive $\epsilon$ prevents an undefined division but cannot recover
information that is absent from the history tensor.

\emph{Proof.}
By definition of RMS for a vector in $\mathbb{R}^{M_lE}$,
\begin{equation}
\left\|\widetilde{\bm{h}}^{(l)}_t\right\|_2^2
=
M_lE\,
\operatorname{RMS}\!\left(
\widetilde{\bm{h}}^{(l)}_t
\right)^2.
\end{equation}
Substitution of Eq.~(\ref{eq:app_rms}) gives
\begin{equation}
\left\|\widetilde{\bm{h}}^{(l)}_t\right\|_2^2
=
M_lE
\left[
r_x\sqrt{\frac{l-1}{M_l}}
\frac{r_h}{r_h+\epsilon}
\right]^2,
\end{equation}
which simplifies to Eq.~(\ref{eq:app_energy_eps}). Setting
$\epsilon=0$ gives Eq.~(\ref{eq:app_energy_exact}). \hfill$\square$

\paragraph{Interpretation.}
The factor $\sqrt{(l-1)/M_l}$ is therefore an energy compensation term.
When pipeline parallelism exposes fewer predecessor layers
($M_l<l-1$), the visible history is amplified so that its total
effective energy remains comparable to a single-stage router with
$l-1$ historical blocks. This does not claim that the individual expert
probabilities become uniform or that routing load is automatically
balanced; those are separate learned properties.

\subsection{Initialization and logit-scale stability}

\paragraph{Proposition 4 (Depth-independent initial logit variance).}
For a layer with $M_l>0$, assume that the entries of the router weight are
independent and sampled from
\begin{equation}
w_{e,i}^{(l)}
\sim
\mathcal{U}(-b_l,b_l),
\qquad
b_l
=
\frac{1}{\sqrt{d+(l-1)E}},
\end{equation}
and that the weights are independent of the router input. Suppose further
that, for some $\sigma_x^2>0$,
\begin{equation}
\mathbb{E}\left[
\left\|\bm{x}^{(l)}_t\right\|_2^2
\right]
=
d\sigma_x^2,
\qquad
\mathbb{E}\left[
\left\|\widetilde{\bm{h}}^{(l)}_t\right\|_2^2
\right]
=
(l-1)E\sigma_x^2.
\label{eq:app_second_moment}
\end{equation}
Then, under the idealized $\epsilon=0$ calibration, every initial router
logit has variance
\begin{equation}
\operatorname{Var}\!\left(
Z_{t,e}^{(l)}
\right)
=
\frac{\sigma_x^2}{3},
\label{eq:app_logit_variance}
\end{equation}
independently of the layer depth $l$ and the number of locally visible
history layers $M_l$.

When $M_l=0$, the historical branch is absent. Under the same
initialization bound, the router receives only the current hidden
representation, and the logit variance becomes
\begin{equation}
\operatorname{Var}\!\left(
Z_{t,e}^{(l)}
\right)
=
\frac{d}{d+(l-1)E}
\frac{\sigma_x^2}{3}.
\label{eq:app_zero_history_variance}
\end{equation}
For the first layer, where $l=1$, this reduces to
$\sigma_x^2/3$.

\emph{Proof.}
Let
\begin{equation}
\bm{u}^{(l)}_t
=
\left[
\bm{x}^{(l)}_t
\middle\|
\widetilde{\bm{h}}^{(l)}_t
\right]
\end{equation}
and let $D_l=d+(l-1)E$. For a uniform random variable on
$[-b_l,b_l]$, the variance is $b_l^2/3$. Since the weights have zero
mean and are independent of the input,
\begin{equation}
\begin{aligned}
\operatorname{Var}\!\left(
Z_{t,e}^{(l)}
\right)
&=
\frac{b_l^2}{3}
\mathbb{E}\left[
\left\|\bm{u}^{(l)}_t\right\|_2^2
\right] \\
&=
\frac{1}{3D_l}
\left(
d\sigma_x^2+(l-1)E\sigma_x^2
\right) \\
&=
\frac{\sigma_x^2}{3}.
\end{aligned}
\end{equation}

For layers with $M_l>0$, the number of locally visible history layers
does not appear in the final expression because
Eq.~(\ref{eq:app_energy_exact}) calibrates the effective historical energy
to $(l-1)E\sigma_x^2$. \hfill$\square$

\paragraph{Remark.}
Proposition 4 is a variance calculation under explicit independence and
second-moment assumptions; it is not a guarantee about the trained
network. When $\epsilon>0$, the historical energy is attenuated by the
factor in Eq.~(\ref{eq:app_energy_eps}). Under the same assumptions, the
historical contribution to the initial router-logit variance is therefore
reduced by the corresponding factor.

\subsection{Branch-wise Initial Logit Variance Decomposition}
\label{app:branch_variance}

The preceding analysis characterizes the variance of the combined router
logits. We further decompose this variance into contributions from the
original current-token branch and the historical residual branch. Consider
a layer with $M_l>0$ and the idealized calibration $\epsilon=0$. For token
$t$ and expert $e$, write
\begin{equation}
Z_{O,t,e}^{(l)}
=
\bm{x}^{(l)}_t
\left(\bm{w}_{O,e}^{(l)}\right)^{\top},
\qquad
Z_{R,t,e}^{(l)}
=
\widetilde{\bm{h}}^{(l)}_t
\left(\bm{w}_{R,e}^{(l)}\right)^{\top},
\end{equation}
so that
\begin{equation}
Z_{t,e}^{(l)}
=
Z_{O,t,e}^{(l)}
+
Z_{R,t,e}^{(l)}.
\end{equation}

\paragraph{Proposition 5 (Branch-wise variance decomposition).}
Under the assumptions of Proposition 4, let
\[
D_l=d+(l-1)E.
\]
Assume that the entries of the original and residual router weight blocks
are independently sampled from $\mathcal{U}(-1/\sqrt{D_l},1/\sqrt{D_l})$.
Then the initial logit variances of the two branches are
\begin{equation}
\operatorname{Var}\!\left(
Z_{O,t,e}^{(l)}
\right)
=
\frac{d}{d+(l-1)E}
\frac{\sigma_x^2}{3},
\label{eq:app_original_branch_variance}
\end{equation}
and
\begin{equation}
\operatorname{Var}\!\left(
Z_{R,t,e}^{(l)}
\right)
=
\frac{(l-1)E}{d+(l-1)E}
\frac{\sigma_x^2}{3}.
\label{eq:app_residual_branch_variance}
\end{equation}
Moreover, the two branches are uncorrelated at initialization:
\begin{equation}
\operatorname{Cov}\!\left(
Z_{O,t,e}^{(l)},
Z_{R,t,e}^{(l)}
\right)
=0.
\label{eq:app_branch_covariance}
\end{equation}
Consequently,
\begin{equation}
\operatorname{Var}\!\left(
Z_{t,e}^{(l)}
\right)
=
\operatorname{Var}\!\left(
Z_{O,t,e}^{(l)}
\right)
+
\operatorname{Var}\!\left(
Z_{R,t,e}^{(l)}
\right)
=
\frac{\sigma_x^2}{3}.
\label{eq:app_total_variance_decomposition}
\end{equation}

\emph{Proof.}
Because the entries of each router weight block have variance
$1/(3D_l)$, the original branch satisfies
\begin{equation}
\begin{aligned}
\operatorname{Var}\!\left(
Z_{O,t,e}^{(l)}
\right)
&=
\frac{1}{3D_l}
\mathbb{E}\!\left[
\left\|
\bm{x}^{(l)}_t
\right\|_2^2
\right] \\
&=
\frac{d\sigma_x^2}{3D_l}
=
\frac{d}{d+(l-1)E}
\frac{\sigma_x^2}{3}.
\end{aligned}
\end{equation}
Similarly, using the idealized energy identity in
Eq.~(\ref{eq:app_energy_exact}),
\begin{equation}
\begin{aligned}
\operatorname{Var}\!\left(
Z_{R,t,e}^{(l)}
\right)
&=
\frac{1}{3D_l}
\mathbb{E}\!\left[
\left\|
\widetilde{\bm{h}}^{(l)}_t
\right\|_2^2
\right] \\
&=
\frac{(l-1)E\sigma_x^2}{3D_l}
=
\frac{(l-1)E}{d+(l-1)E}
\frac{\sigma_x^2}{3}.
\end{aligned}
\end{equation}
The covariance is zero because the two weight blocks are independently
sampled and have zero mean. Adding the two branch variances gives
Eq.~(\ref{eq:app_total_variance_decomposition}). \hfill$\square$

\paragraph{Interpretation.}
The scale-preserving initialization keeps the total initial router-logit
variance independent of layer depth, but it does not force the original and
historical branches to have equal contributions. Their idealized variance
fractions are
\begin{equation}
\eta_O^{(l)}
=
\frac{d}{d+(l-1)E},
\qquad
\eta_R^{(l)}
=
\frac{(l-1)E}{d+(l-1)E},
\qquad
\eta_O^{(l)}+\eta_R^{(l)}=1.
\end{equation}
Thus, under the global fan-in used by HERO-MoE, the relative contribution of
the historical branch increases with depth while the total logit scale
remains controlled. The number of locally visible history layers $M_l$ does
not appear in these idealized variance fractions because the
scale-preserving factor compensates for the corresponding historical
energy. With finite $\epsilon>0$, the historical contribution is attenuated
in a token-dependent manner, and the exact residual-branch variance
depends on the distribution of $r_h$.

\subsection{Theoretical implications}
The preceding propositions establish the following properties of
HERO-MoE:
\begin{enumerate}
\item The proposed router is a strict extension of the baseline router:
      removing historical information recovers the original routing
      function exactly.
\item Historical routing priors provide an additional decision signal
      without creating an explicit gradient path through earlier routing
      distributions.
\item For layers with $M_l>0$, the factor $\sqrt{(l-1)/M_l}$ compensates for the number of visible historical layers, making the calibrated effective historical energy independent of $M_l$ in the idealized $\epsilon=0$ setting.
\item The historical branch modifies only the router logits. Therefore,
      any standard sparse routing operator can be applied after the
      proposed router without changing its definition.
\end{enumerate}
Together, these results provide a theoretical explanation for the three
design goals of HERO-MoE: compatibility with standard sparse MoE
routing, decoupling of historical priors from earlier router gradients,
and calibrated historical signal magnitude across depth and different numbers of visible historical layers.

\SetCommentSty{textrm}

\begin{algorithm}[t]
\scriptsize
    \caption{Depth-aware Initialization of the HERO-MoE Router}
    \label{alg:hero-init}
    \SetKwInOut{Input}{Input}
    \SetKwInOut{Output}{Output}
    \Input{
        Total number of MoE layers $L$, 
        current layer index $l$, 
        hidden dimension $d$, 
        number of experts $E$, 
        pipeline-stage assignment $\pi(\cdot)$
    }
    \Output{
        Router weight $\bm{W}^{(l)}$
    }
    \tcp{1. Determine the visible historical layers}
    $\mathcal{S}(l) \leftarrow
    \{j \mid j<l,\ \pi(j)=\pi(l)\}$\;
    $M_l \leftarrow |\mathcal{S}(l)|$\;
    \tcp{2. Determine the physical router dimension}
    $D_l \leftarrow d + M_l E$\;
    \tcp{$D_l=d$ when no preceding layer is visible}
    \tcp{3. Use the global depth to determine the initialization scale}
    $b_l \leftarrow \dfrac{1}{\sqrt{d+(l-1)E}}$\;
    \tcp{4. Initialize the single concatenated router matrix}
    Initialize each entry
    $W^{(l)}_{e,i}\sim\mathcal{U}(-b_l,b_l)$
    for $e\in\{1,\ldots,E\}$ and $i\in\{1,\ldots,D_l\}$\;
    \eIf{$M_l=0$}{
        $\bm{W}_{O}^{(l)} \leftarrow \bm{W}^{(l)}$\;
        $\bm{W}_{R}^{(l)} \leftarrow \emptyset$\;
    }{
        Partition $\bm{W}^{(l)}$ as
        $\bm{W}^{(l)}
        =
        \left[
        \bm{W}_{O}^{(l)}
        \middle\|
        \bm{W}_{R}^{(l)}
        \right]$\;
        $\bm{W}_{O}^{(l)}\in\mathbb{R}^{E\times d}$,\quad
        $\bm{W}_{R}^{(l)}\in\mathbb{R}^{E\times M_lE}$\;
    }
    \Return{$\bm{W}^{(l)}$}
\end{algorithm}

\section{Algorithm of HERO-MoE}
\label{sec:appendix_algorithm}
The pseudocode of HERO-MoE is divided into two procedures.
Algorithm~\ref{alg:hero-init} initializes the history-aware router with a
depth-aware global fan-in. Algorithm~\ref{alg:hero-routing} performs
stage-local history retrieval, scale-preserving fusion, router-logit
computation, and standard sparse expert selection.

\begin{algorithm}[t]
\scriptsize
    \caption{Stage-local Historical Routing in HERO-MoE}
    \label{alg:hero-routing}
    \SetKwInOut{Input}{Input}
    \SetKwInOut{Output}{Output}
    \SetKwFunction{softmax}{softmax}
    \SetKwFunction{RMS}{RMS}
    \SetKwFunction{StopGrad}{StopGrad}
    \SetKwFunction{SparseRoute}{SparseRoute}
    \Input{
        Current hidden states $\bm{X}^{(l)}\in\mathbb{R}^{N\times d}$,
        current layer index $l$,
        pipeline-stage assignment $\pi(\cdot)$,
        preceding router logits $\{\bm{Z}^{(j)}\}_{j<l}$,
        router weight $\bm{W}^{(l)}$,
        numerical constant $\epsilon>0$,
        sparse routing configuration $\mathcal{C}$
    }
    \Output{
        Router logits $\bm{Z}^{(l)}$,
        routing weights $\bm{G}^{(l)}$,
        selected expert indices $\bm{I}^{(l)}$,
        dense routing distribution $\bm{Q}^{(l)}$
    }
    \tcp{1. Identify stage-local routing history}
    $\mathcal{S}(l) \leftarrow
    \{j \mid j<l,\ \pi(j)=\pi(l)\}$\;
    $M_l \leftarrow |\mathcal{S}(l)|$\;
    \eIf{$M_l=0$}{
        $\bm{U}^{(l)} \leftarrow \bm{X}^{(l)}$\;
    }{
        \tcp{2. Collect detached dense routing distributions}
        Enumerate
        $\mathcal{S}(l)=\{j_1,j_2,\ldots,j_{M_l}\}$\;
        \For{$m\leftarrow 1$ \KwTo $M_l$}{
            $\bm{Q}^{(j_m)}
            \leftarrow
            \StopGrad\!\left(
            \softmax\!\left(\bm{Z}^{(j_m)}\right)
            \right)$\;
        }
        $\bm{H}^{(l)}
        \leftarrow
        \left[
        \bm{Q}^{(j_1)},
        \bm{Q}^{(j_2)},
        \ldots,
        \bm{Q}^{(j_{M_l})}
        \right]$\;
        \tcp{3. Apply token-wise scale-preserving fusion}
        $\bm{r}_{x}
        \leftarrow
        \RMS\!\left(\bm{X}^{(l)}\right)$\;
        $\bm{r}_{h}
        \leftarrow
        \RMS\!\left(\bm{H}^{(l)}\right)$\;
        $\bm{\rho}^{(l)}
        \leftarrow
        \dfrac{\bm{r}_{x}}{\bm{r}_{h}+\epsilon}
        \sqrt{\dfrac{l-1}{M_l}}$\;
        $\widetilde{\bm{H}}^{(l)}
        \leftarrow
        \bm{\rho}^{(l)}\odot\bm{H}^{(l)}$\;
        \tcp{4. Concatenate current features and historical features}
        $\bm{U}^{(l)}
        \leftarrow
        \left[
        \bm{X}^{(l)}
        \middle\|
        \widetilde{\bm{H}}^{(l)}
        \right]$\;
    }
    \tcp{5. Compute history-aware router logits}
    $\bm{Z}^{(l)}
    \leftarrow
    \bm{U}^{(l)}\bm{W}^{(l)\top}$\;
    \tcp{6. Form dense routing probabilities for the current layer}
    $\bm{Q}^{(l)}
    \leftarrow
    \softmax\!\left(\bm{Z}^{(l)}\right)$\;
    \tcp{7. Apply the configured sparse routing policy}
    $(\bm{G}^{(l)},\bm{I}^{(l)})
    \leftarrow
    \SparseRoute\!\left(
    \bm{Z}^{(l)},\bm{Q}^{(l)},\mathcal{C}
    \right)$\;
    \tcp{$\mathcal{C}$ may specify top-$k$, group-limited routing,
    capacity, and top-$k$ weight normalization}
    \Return{$
    \bm{Z}^{(l)},\ 
    \bm{G}^{(l)},\ 
    \bm{I}^{(l)},\ 
    \bm{Q}^{(l)}
    $}
\end{algorithm}

\section{Visualization and Statistical Details}
\label{app:visualization_details}

This section describes the data processing and statistics used to generate
Figures~\ref{fig:routing_analysis}.

\paragraph{Layer-wise dependency and expert-level coupling.}
The routing analysis uses the trained router weights of a 30-layer MoE model
with $E=128$ experts and hidden dimension $d=1024$. The model is partitioned
into two pipeline-parallel stages, with 15 layers per stage. The implementation
uses zero-based layer indices $0,\ldots,29$, whereas the equations in the main
text use one-based layer indices. Thus, an implementation layer index $l_0$
corresponds to the mathematical layer index $l=l_0+1$.

For a target layer $l$ and a preceding layer $p$ located in the same
pipeline-parallel stage, let
\[
\bm{W}^{(l,p)}_{\mathrm{R}}\in\mathbb{R}^{E\times E}
\]
denote the residual-router weight block associated with the routing history
from layer $p$ to layer $l$. For each visible predecessor, the block is
extracted from the columns following the first $d$ current-token features.
The layer-wise dependency strength shown in
Figure~\ref{fig:routing_analysis}(a) is computed as
\begin{equation}
D_{l,p}
=
\left\|\bm{W}^{(l,p)}_{\mathrm{R}}\right\|_{\mathrm{F}}
\sqrt{\frac{l-1}{M_l}},
\qquad
p<l,
\end{equation}
where $M_l$ is the number of preceding layers visible on the same
pipeline-parallel stage. The multiplier is the implementation equivalent of
the depth-compensation factor in Eq.~(\ref{eq:rou}); it is computed as
$\sqrt{l_0/M_l}$ using the zero-based layer index $l_0$. Entries
corresponding to cross-stage layer pairs are set to zero because HERO-MoE
uses stage-local routing history. Therefore, Figure~\ref{fig:routing_analysis}(a)
contains two independent $15\times15$ diagonal blocks.

To quantify how strong expert-level interactions vary with layer distance, we
define
\begin{equation}
\mathcal{W}_{\Delta}
=
\left\{
\left(\bm{W}^{(l,p)}_{\mathrm{R}}\right)_{a,b}
:
l-p=\Delta,\;
p\text{ and }l\text{ are in the same stage}
\right\},
\end{equation}
where $\Delta\in\{1,\ldots,14\}$ and $a,b\in\{1,\ldots,E\}$. A single global
threshold is computed from all historical-router entries:
\begin{equation}
\tau
=
\operatorname{Percentile}_{90}
\left(
\left\{
|w|:
w\in\bigcup_{\Delta=1}^{14}\mathcal{W}_{\Delta}
\right\}
\right).
\end{equation}
For each inter-layer distance $\Delta$, the negative and positive strong-coupling
ratios are then calculated as
\begin{equation}
r^{-}_{\Delta}
=
\frac{100}{|\mathcal{W}_{\Delta}|}
\sum_{w\in\mathcal{W}_{\Delta}}
\mathbb{1}[w<-\tau],
\qquad
r^{+}_{\Delta}
=
\frac{100}{|\mathcal{W}_{\Delta}|}
\sum_{w\in\mathcal{W}_{\Delta}}
\mathbb{1}[w>\tau].
\end{equation}
Figure~\ref{fig:routing_analysis}(b) visualizes $r^{-}_{\Delta}$ and
$r^{+}_{\Delta}$ as stacked bars. Thus, this panel reports the fraction of
strongly negative or positive coupling entries among all expert-pair weights
at a given layer distance, rather than the fraction of activated experts.

Finally, Figure~\ref{fig:routing_analysis}(c) visualizes the raw residual
weight block for the implementation pair Layer 8 $\rightarrow$ Layer 9.
The horizontal axis corresponds to source experts, and the vertical axis
corresponds to target experts. The color range is symmetric around zero and
is clipped at the 99.8th percentile of the absolute matrix values to reduce
the influence of extreme entries. Positive and negative colors therefore
indicate the signs of learned residual-router coefficients. These signs serve
as a weight-based structural proxy for positive and negative routing influence,
rather than a direct causal measurement of expert cooperation or competition.

\end{document}